%% file: lnas.tex
\documentclass{article} 
\usepackage{iclr2021_conference,times}

\input{math_commands.tex}

\usepackage{float}
\usepackage{hyperref}
\usepackage{url}
\usepackage{graphicx}
\usepackage{subfig}

\restylefloat{table}

\title{Self-Constructing Neural Networks through Random Mutation}

\author{Samuel Schmidgall \\
Department of Computer Science\\
George Mason University\\
\texttt{sschmidg@gmu.edu}
}

\iclrfinalcopy 
\begin{document}

\maketitle

\begin{abstract}
The search for neural architecture is producing many of the most exciting results in artificial intelligence. It has increasingly become apparent that task-specific neural architecture plays a crucial role for effectively solving problems. This paper presents a simple method for learning neural architecture through random mutation. This method demonstrates 1) neural architecture may be learned during the agent's lifetime, 2) neural architecture may be constructed over a single lifetime without any initial connections or neurons, and 3) architectural modifications enable rapid adaptation to dynamic and novel task scenarios. Starting without any neurons or connections, this method constructs a neural architecture capable of high-performance on several tasks. The lifelong learning capabilities of this method are demonstrated in an environment without episodic resets, even learning with constantly changing morphology, limb disablement, and changing task goals all without losing locomotion capabilities. 
\end{abstract}

\section{Introduction}



Biological brains learn as a product of synaptic and structural changes that occur over a single continuous lifetime. While structural developments in the brain are largely tailored by genetic factors, many of the genetic developments are dependent on experience for proper development (\cite{berardi2015brain}). Early experimental evidence for this was shown in (\cite{cat_depr}) where cats were blindfolded in one eye during a critical developmental period. As a result, development of the neurons and neuronal connectivity responsible for visual processing was significantly favored in the unimpaired eye. This type of structural plasticity is not limited to critical developmental periods, and generally occurs throughout the lifetime of mature organisms (\cite{struct_plast}, \cite{Holtmaat2009}, \cite{JOHANSENBERG2007R141}). 

In nature, animals must learn without episodic boundaries or "environment resets," which is referred to as lifelong learning. When an animal makes a mistake it must face the consequences and be capable of learning additional skills to correct itself. This differs from the typical Reinforcement Learning (RL) setting, where the agent's lifetime is composed of the same task repeated over a small time interval. Often in RL, if a mistake is made, the simulation just starts over again, side-stepping the need to learn a more general set of skills. Part of the difficulty with lifelong learning settings is that many learning algorithms experience a phenomenon known as catastrophic forgetting, which causes previously learned capabilities to degrade with the incorporation of new knowledge. Many successful attempts to overcome catastrophic forgetting involve localizing learning to a subset of the existing neural connections (\cite{ewcs}, \cite{masse2018alleviating}). However, this often requires knowledge of a discrete set of different tasks, as well as a centralized mechanism that switches between these regions in the right context. Additionally, it requires pre-defining the number of connections, hence the amount of learning that is expected to take place must be known prior to training. More often than not interesting learning tasks are not so simple. 


In order to learn without episodic boundaries while constantly constructing new neurons and neural connections, a simple method for actively modifying neural architecture is proposed. This is accomplished by randomly modifying the neural architecture and incorporating changes based on a dynamic acceptance threshold. This method is demonstrated to be capable of lifelong learning in a constantly changing environment, constructing networks capable of learning with only a small set of weight values, and performing competitively with traditional learning methods despite its simplicity. 







\section{Related Work}

While active self-construction itself is a less explored topic, the development of neural architecture emerged from a long history in evolutionary computation (\cite{neat}, \cite{risi2011enhancing}, \cite{stanley2009hypercube}), and has had success more recently with random search (\cite{rand_search}, \cite{searchphase}) and gradient-based methods (\cite{zoph2017neural}, \cite{hypernetworks}, \cite{li2020geometryaware}). The goal of Neural Architecture Search (NAS) is to automatically find an architecture that, once trained, outperforms hand-designed architectures (\cite{elsken2019neural}). As a consequence, many of these methods involve an expensive inner and outer loop search of optimizing architecture and weights. A classic example of architecture search which optimizes both weights and topology simultaneously is NeuroEvolution of Augmenting Topologies (NEAT) (\cite{neat}). NEAT begins with a population of minimal neural network topologies and increasingly leads to a more complex and diverse population. These changes are made through random mutation on the evolutionary scale across generations.

A related strategy for NAS, network pruning, strategically removes connections from densely connected neural networks once their weight value shrinks below a certain value. This method has been shown to produce networks that contain only a small fraction of their original number of connections, while still retaining competitive performance with its fully-connected counterpart on the task they were trained for (\cite{kusupati2020soft}, \cite{dong2019network}). Some results even indicate removing redundant connections may improve performance (\cite{DBLP:journals/corr/abs-1912-04427}, \cite{frankle2018lottery}). Recent work considered the capabilities of networks that were \textit{only pruned}, using randomly initialized weights which were pruned down without weight optimization to produce networks that had far better than chance performance on several image classification tasks (\cite{rand_prune}).


The importance of architecture itself independent of weight values was first explored in (\cite{wann2019}). Here, a population of neural architectures are evolved using NEAT; however, to evaluate the population independent of weights, the performance of an architecture is collected using random weights. Since sufficient performance could not be attained using completely random weights, the performance of these networks are evaluated over a single weight which is shared among every connection. The overall performance of an architecture is computed by averaging performance over several of these shared weights taken from a discrete set. In this work, weight-agnostic neural networks were shown to provide far better than chance performance on several RL tasks indicating that neural architecture itself is may provide a solution to the learning problem.




\section{Random Mutation Search}

Random Mutation Search (RMS) may be summarized in a few simple steps: \textbf{(1)} Initialize and evaluate a network of input and output neurons with no connections and set as current best, \textbf{(2)} randomly mutate the current best network structure, \textbf{(3)} evaluate network performance over a specified time interval, \textbf{(4)} if the observed performance is better than the previous best network (mutation acceptance threshold) then update best network and performance, \textbf{(5)} repeat algorithm starting at step \textbf{2}.

\begin{figure}
\centering
\includegraphics[width=13.6cm]{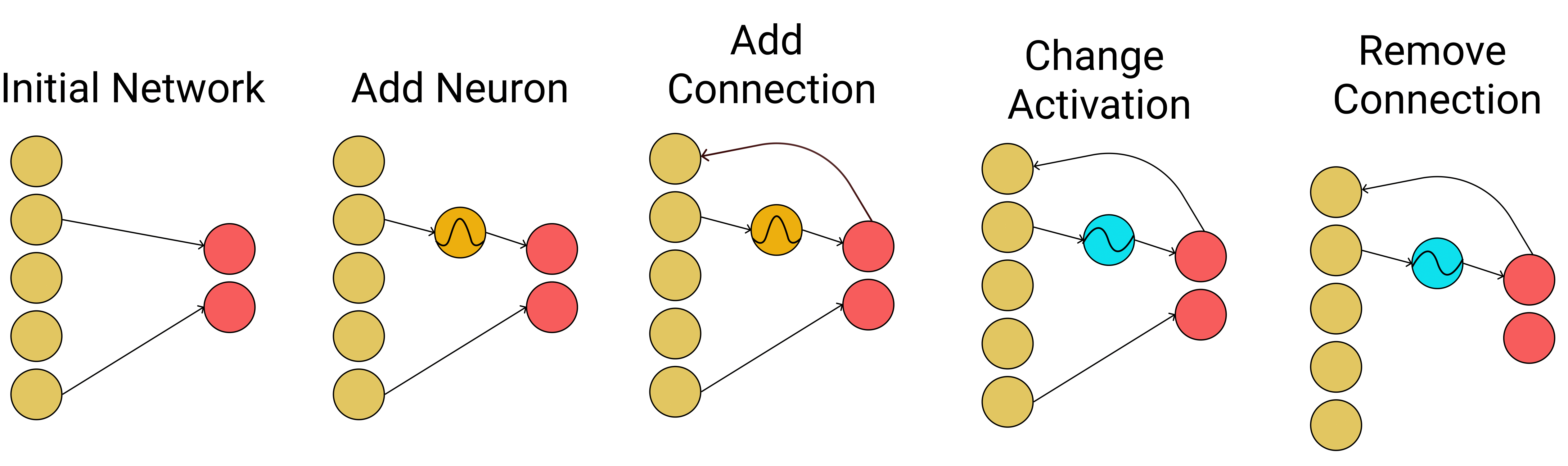}
\caption{A network with five input neurons (\textit{yellow}) and two action neurons (\textit{red}) initialized with two connections. The network is altered with \textit{Add Neuron}, splitting a random connection and adding a neuron with a random activation in-between. \textit{Add Connection} randomly connects two neurons. \textit{Change Activation} selects a neuron and randomly chooses a new activation. \textit{Remove Connection} removes a randomly selected connection.}
\label{operators}
\end{figure}

The mutation operator has four mutation types: add connection, add neuron, change neuron activation, and remove connection (Figure \ref{operators}). These operations provide the capability for both neural pruning and topology augmentation. The network is represented as a directed graph structure. This graph structure supports the development of feedforward, feedback, and self-recurrent connections. While feedforward and self-recurrent weights are commonly incorporated in neural networks, feedback weights are less utilized largely because it is unknown how and when to introduce these weights manually (\cite{feedbackgood}). 

\textbf{Add Connection}: The add connection operator randomly selects two neurons and adds a connection between them. The connection's weight is then assigned to a random value. This operation allows for any type of connection including: feedforward, feedback, lateral, and self-recurrent. Feedback weights connect one neuron with another whose activation occurs before it. Similar to self-recurrence, feedback connections receive a delayed signal from the activated neuron. Lateral connections occur between two neurons in the same layer. While the "hidden" network structure does not have a strict definition, lateral connections are more straightforwardly seen as connecting two output or input neurons and are treated in the same way as feedback connections with a delayed signal. 

\textbf{Add Neuron}: To add a neuron, a connection is selected at random, and split into two connections with a new neuron inbetween. As in (\cite{neat}), the weight of the new neuron's incoming connection is set to 1, and the weight of the connection that was split is inherited by the new neuron's outgoing connection. The new neuron's activation function is then randomly selected.

\textbf{Change Activation}: The activation change operator is the most straightforward. A neuron is selected at random and its activation is randomly sampled from a list of defined activations. In this work a list of common (e.g. ReLU, hyperbolic tangent, sigmoid) and uncommon activation functions (e.g. Gaussian, sinusoid, step) following from (\cite{wann2019}) are selected.


\textbf{Remove Connection}: Like the add connection operator, a connection is selected at random; however, actually removing the connection comes with additional difficulties. Once the connection is removed, the receiving neuron may end up without incoming connections. Additionally, the sending neuron may no longer have any outgoing connections. In either of these cases, a neuron deletion strategy must be employed. There are four primary strategies for deciding when to perform neuron deletion: when all incoming connections have been removed, when all outgoing connections have been removed, when both sets have, or when either set has been removed. In all but the final scenario, neurons and connections may exist in the networks without actually contributing to neural activity. However, these three methods preserve neural connections that were at one point useful which may be reused at a later time.



\section{Experiments}

\captionsetup[subfigure]{labelformat=empty}

The experimental section evaluates two distinct learning settings to provide an intuition for the expected performance, capabilities, and behavior of these self-constructing networks. In the performance comparison section, RMS is compared with Evolutionary Strategies (\cite{salimans2017evolution}), NEAT (\cite{neat}), and the on-policy gradient method Proximal Policy Optimization (\cite{ppo}) on several RL tasks in the traditional episodic setting. The lifelong learning section applies RMS in an environment without episodic resets all while constantly experiencing changes in morphology, limb impairment, and task goal.


For all of the experiments in this paper, neuron deletion occurs when both sets of incoming and outgoing connections have been removed. Upon creation, connection weights are randomly assigned to values in the set $\{\pm1, \pm0.5, \pm0.25\}$, which is referred to as the discrete set. In addition to the original four operators, a fifth is added, \textit{change weight}, which randomly assigns a connection weight to a value sampled from the discrete set. In more practical settings this operation would be replaced by weight optimization techniques such as reinforcement learning, local plasticity rules, etc. To account for performance evaluation variance, as well as the changing environmental dynamics observed during lifetime learning problems, the current best network performance discussed in \textbf{(4)} is decayed exponentially over time. Additionally, in the episodic setting, RMS is penalized for having high variance among collected episodes; hence, the mutation acceptance threshold increases for that individual mutation to promote consistent behavior.



\subsection{Comparison to other methods}

The performance of three alternate learning paradigms are compared with RMS in the traditional episodic learning setting. The purpose of this section is provide a frame of reference for the learning capabilities of RMS. The first method, NEAT, provides the most straightforward comparison since the neural architecture does not need to be selected prior to training; it is constructed from scratch. The Evolutionary Strategies (ES) optimization, like NEAT, is a population based optimization method. However, unlike NEAT, ES does not construct its own topology; rather it is a numerical optimizer which follows from (\cite{salimans2017evolution}). As a result, the network topology must be hand-designed before training time. The final method uses stochastic gradient ascent via the on-policy RL algorithm Proximal Policy Optimization (PPO) (\cite{ppo}). Like ES, this method is not self-constructing. The hyperparameters for each of these methods are described in the Appendix\footnote{The code is available at https://github.com/SamuelSchmidgall/RandomMutationSearch}.

\begin{table}[H]
    \centering
    \begin{tabular}{||c|c|c|c||}
    \hline
        Optimization Method & Ant Locomotion & Swimmer & Swing-up\\
        \hline
        \hline
        Random Mutation Search & \textbf{1076 $\pm$ 106} & 1106 $\pm$ 39 & 326 $\pm$ 35 \\ 
        \hline 
        NEAT & 788 $\pm$ 55 & 1177 $\pm$ 59 & 363 $\pm$ 14\\
        \hline 
        Evolutionary Strategies & 737 $\pm$ 113 & \textbf{1318 $\pm$ 12} & 218 $\pm$ 3\\
        \hline
        Proximal Policy Optimization & 962 $\pm$ 17 & 1286 $\pm$ 31 & \textbf{379 $\pm$ 12}\\
        \hline 
    \end{tabular}
    \caption{Performance of Random Mutation Search compared with three other algorithms on robotic learning benchmarks. Descriptions of these tasks are provided in the Appendix.}
    \label{tab:performance}
\end{table}

The performance results of each learning strategy is described in (Table \ref{tab:performance}). Despite having only a discrete set of possible weight values, and making changes based on randomly selected mutations, RMS is capable of performing competitively with ES, NEAT, and PPO.

For the Ant Locomotion task, RMS developed a total average of 283 connections (Table \ref{tab:connections}) which was 95\% less than the 6336 connections that were used in PPO and ES, and 28\% less than NEAT's 393 connections. The ratio of feedback, feedforward, and self-recurrent weights were relatively consistent across tasks even as the number of connections increased. Self-recurrent connections did not develop frequently since sampling the same neuron in the \textit{add connection} operation became less likely as more neurons were added. Comparing the difficulty of a task is challenging across different learning settings (e.g. Ant Locomotion, Swimmer, Swing-up), hence making it difficult to know the effect of task complexity on learned architecture. To appropriately assess this effect, two additional Ant Locomotion settings are considered of varying complexity. The limb impairment setting has the same goal as Ant Locomotion; however, a limb is randomly selected at the beginning of each episode, and all action output to that limb is rendered void (Figure \ref{ant} (A)). The dynamic morphology setting randomizes limb length, weight, and width (Figure \ref{ant} (B)). The results of training Ant Locomotion in these two settings, as well as in the non-impaired setting, indicate an increase in the number of connections across tasks based on task difficulty. The limb impaired architecture had 7\% more connections than the non-impaired, and the dynamic morphology had 27\% more. As a product of RMS co-developing weights and architecture without populations, it only required 10 million environment interactions compared to NEAT's 150 million.

\begin{table}[H]
    \centering
    \begin{tabular}{||c|c|c|c|c||}
    \hline
        Quadruped Task & Connections & Feedforward & Feedback & Self-Recurrent\\
        \hline
        \hline
        Locomotion Only & 283 & 94\% & 6\% & 0\%\\ 
        \hline
        Limb Impairment & 302 & 92\% & 8\% & 0\%\\ 
        \hline
        Dynamic Morphology & 362 & 92\% & 7\% & 1\%\\ 
        \hline
        Lifelong & 483 & 93\% & 7\% & 0\%\\ 
        \hline
    \end{tabular}
    \caption{Average composition of learned neural architectures by connection type on Ant locomotion task over 10 million environment interactions in increasing order of connections.}
    \label{tab:connections}
\end{table}

\subsection{Lifelong Learning}

In the lifelong learning scenario, a variation of Ant locomotion is introduced in a setting where episodic boundaries are dissolved. As a consequence, the learning problem changes fundamentally. In the episodic setting of this task, when the quadruped falls over, becomes disoriented, or positions its limbs poorly there are few consequences, as the episodic nature of the task simply restarts the quadruped's position. This results in minimal incentive for learning a more general task solving skill-set. Additionally, these tasks are simulated over a very short time-period which is comparatively only a small fraction of the lifespan of even the simplest creatures on earth. There is a significant difference in the way biological organisms must learn compared to traditional simulated machines.

\begin{figure}[H]
\centering
\subfloat[\centering (A) Impaired Limb]{{\includegraphics[width=5cm]{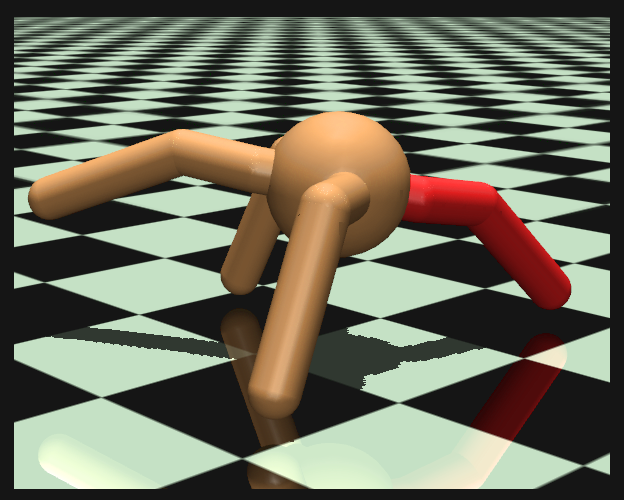} }}%
\qquad
\subfloat[\centering (B) Randomized Morphology]{{\includegraphics[width=6.765cm]{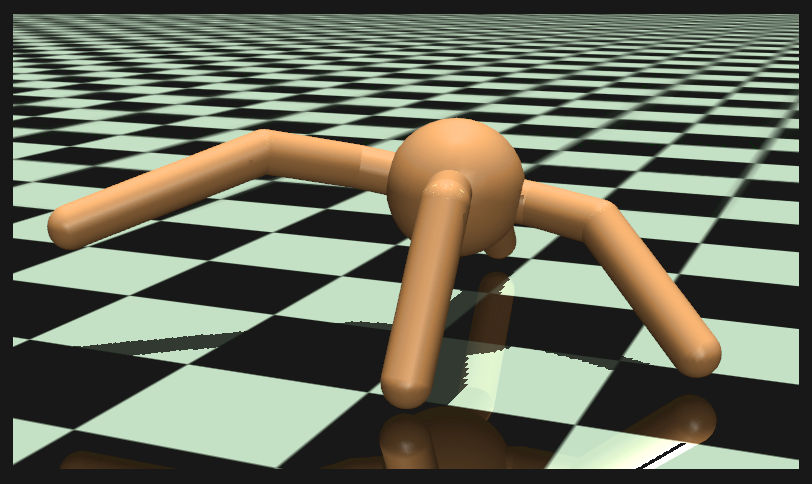} }}%
\caption{Variations of the original Ant locomotion task encountered in the lifelong learning setting. (A) A random limb is chosen to be impaired \textit{(in red)} rendering all action output to that limb void. (B) In the dynamic morphology task variant, morphological changes are randomly applied causing differences in limb length, width, and weight.}
\label{ant}
\end{figure}


An important component of lifelong learning is that the task itself requires adaptation to new scenarios from which general knowledge can be developed. Here, this is introduced into the locomotion task in three ways: 1) the quadruped's body is constantly changing (Figure \ref{ant}B), 2) the quadruped's limbs are constantly being impaired and unimpaired (Figure \ref{ant}A), 3) the target direction for locomotion is constantly changing at random intervals all without environment resets. Importantly, the morphological changes are not incorporated into the state information, hence the self-constructing network must learn based on implicit context signals which limb is impaired and how the body was changed as well as produce a more general pattern of locomotion to account for these changes.

The lifelong learning networks were trained over 10 million environment interactions, just as the episodic networks were in (Section 4.1). However, self-construction led to the development of a network that was $71\%$ larger than in the episodic setting, once again indicating there is an increase in network size based on task complexity.

The performance metric for this task is different than the episodic Ant Locomotion task to account for direction heading; however, to evaluate the efficacy and generalization of the lifelong learning neural architecture, we apply it to the Ant Locomotion task without dynamic morphology. Overall, there was little difference between the performance of the network trained in the episodic setting and the lifelong learning setting despite having been trained for different tasks; the lifelong learning network obtained a mean score of \textbf{856} compared to the episodic \textbf{1076}. The similarity is non-trivial since generalization, in this case over morphology, impairment, and direction, is often accompanied by a decrease in specialization. There was only a slight decrease in performance for three of the four limbs that were disabled individually: the two front limbs obtained scores of \textbf{822}, \textbf{789} and the back left limb, \textbf{658}. Interestingly, with the back right limb disabled the policy obtained a low performance average of \textbf{22}. This is because the self-construction mechanism generated a more \textit{greedy} solution\footnote{This strategy was visually reminiscent of cellular movement using a flagella, with the remaining limbs providing stability in the front.} to provide limb robustness by, instead of relying on all four limbs equally, putting all of the locomotion capabilities into a single limb so that there was a lower probability of performance impairment. This is likely a result of selecting architecture based purely on maximum performance in a setting where the limb impairment is randomly sampled; episodes where the dominant limb is not disabled are likely to be preferred over those where it is.

There was little effect on morphological randomization and performance, obtaining an average of \textbf{844}. Since the lifelong neural architecture provides similar performance despite the necessity of generalization, perhaps an enlargement of neural structure may provide both general and specialized learning capabilities.


\section{Discussion}

Despite RMS proving successful on the task set presented in this work, environments with large observation or action spaces may likely prove challenging for RMS to effectively generate meaningful connections using only random mutations. This was a central concern around NEAT strategies. A potential alleviation of this challenge may come from using RMS to develop a network that serves to generate a separate network which acts as the environment policy, as in (\cite{stanley2009hypercube}, \cite{hypernetworks}). Additional performance improvements may come from combining weight-optimization techniques together with RMS, and could serve as a simple mechanism for both architectural and weight optimization that is capable of developing in lifelong learning settings. 

This work introduces a simple method for learning neural architecture through random mutation. As a product of its simplicity, this method is capable of learning architecture during the agent's lifetime. This innovation was shown to provide competitive performance with several standard learning paradigms, as well as being applied to a lifelong learning setting where episodic boundaries are dissolved and dynamic changes in morphology and task setting are constantly being experienced. Despite this, single weight optimizations were a product of random mutations sampled from a small discrete set of possible values. The learned architectures were marginally smaller than those developed by NEAT, and significantly smaller than the standard static architectures used in PPO and ES. In the lifelong learning setting, as well as the limb impairment and dynamic morphology settings, the number of neural connections were observed to increase as a product of task complexity.







\bibliography{lnas}
\bibliographystyle{iclr2021_conference}

\appendix

\section{Training Details}

\subsection{Hyperparameters}

This section provides an in-depth set of hyperparameter tables for reproduciblity, which is also \hyperlink{https://github.com/SamuelSchmidgall/RandomMutationSearch}{\textit{\textbf{available in the code}}} (\cite{random_mutation_search}). The code to train PPO in the performance comparison section was taken from OpenAI Baselines (\cite{baselines}).

\begin{table}[H]
    \centering
    \begin{tabular}{||c|c||}
    \hline
    \multicolumn{2}{|c|}{RMS Hyperparameters}\\
    \hline
    \hline
    Mutation per Update (Ant, Swimmer, Lifelong) & $\mathbb{U}(1, 20)$ \\
    \hline
    Mutation per Update (Swingup) & $1$ \\
    \hline
    Connection Penalty & 0.001  \\ 
    \hline
    Top Performance Decay Rate & 0.995  \\ 
    \hline
    Initial Random Mutations & 50  \\ 
    \hline
    Weight Type & $\{\pm1, \pm0.5, \pm0.25\}$  \\ 
    \hline
    Neuron-deletion Strategy & Both-sets  \\ 
    \hline 
    Remove Connection Probability & 0.2  \\ 
    \hline
    Add Connection Probability & 0.2  \\ 
    \hline
    Change Activation Probability & 0.2  \\ 
    \hline
    Add Neuron Probability & 0.2  \\ 
    \hline
    Change Weight Probability & 0.2  \\ 
    \hline
    Performance Variance Penalty Coeff. & 0.1 \\
    \hline
    Update Interval & 4000  \\  
    \hline
    \end{tabular}
    \label{tab:hyperparamtableRMS}
\end{table}

\begin{table}[H]
    \centering
    \begin{tabular}{||c|c||}
    \hline
    \multicolumn{2}{|c|}{ES Hyperparameters}\\
    \hline
    \hline
    Population Size & 128 \\
    \hline
    Learning Rate & 0.04  \\  
    \hline
    Learning Rate Decay & 0.999  \\      
    \hline
    Parameter Sampling Type & Normal  \\  
    \hline
    Parameter Std & 0.04  \\  
    \hline
    Parameter Std Decay & 0.999  \\  
    \hline
    Weight Penalty & 0.01  \\  
    \hline
    Hidden Layer Width & 64  \\  
    \hline
    Hidden Layers & 2  \\  
    \hline
    Hidden Neuron Activation & $tanh$  \\  
    \hline
    Action Neuron Activation & $tanh$  \\  
    \hline
    Action Noise Std & 0.01  \\  
    \hline
    Update Iterations & 512  \\  
    \hline
    \end{tabular}
    \label{tab:hyperparamtableES}
\end{table}

\begin{table}[H]
    \centering
    \begin{tabular}{||c|c||}
    \hline
    \multicolumn{2}{|c|}{NEAT Hyperparameters}\\
    \hline
    \hline
    Population Size & 192 \\
    \hline
    Parameter Sampling Type & Normal  \\  
    \hline
    Parameter Standard Deviation & 0.005  \\  
    \hline
    Optimization Type & Elite Selection  \\      
    \hline
    Total Elites & 24  \\  
    \hline
    Unchanged Elites & 2  \\  
    \hline
    Mutations Per Update & 1  \\  
    \hline
    Add Neuron Probability & 0.2  \\  
    \hline
    Add Connection Probability & 0.4  \\  
    \hline
    Change Activation Probability & 0.4  \\ 
    \hline
    Performance Episode Average & 3  \\ 
    \hline
    Update Iterations & 512  \\  
    \hline
    \end{tabular}
    \label{tab:hyperparamtableNEAT}
\end{table}

\begin{table}[H]
    \centering
    \begin{tabular}{||c|c||}
    \hline
    \multicolumn{2}{|c|}{PPO Hyperparameters}\\
    \hline
    \hline
    PPO Gamma ($\gamma$) & 0.99  \\  
    \hline
    PPO Lambda ($\lambda$) & 0.95  \\       
    \hline
    Clip Parameter & 0.2  \\  
    \hline
    Optimizer Step Size & 3e-4  \\  
    \hline
    Actor Critic Batch Size & 2048  \\  
    \hline
    Optimizer Batch Size & 64  \\  
    \hline
    Batch Epochs & 10  \\  
    \hline
    Hidden Layer Width & 64  \\  
    \hline
    Hidden Layers & 2  \\  
    \hline
    Hidden Neuron Activation & $tanh$  \\  
    \hline
    Action Neuron Activation & $linear$  \\  
    \hline
    Entropy Coeff. & 0.0  \\  
    \hline
    Total Interactions & 1 Million  \\  
    \hline
    Optimizer & Adam (\cite{kingma2014adam})  \\ 
    \hline
    \end{tabular}
    \label{tab:hyperparamtablePPO}
\end{table}
















\subsection{Environment Descriptions}

Here a set of descriptions are provided for each environment used in this work. We note that these environments are also available in the paper code repository (\cite{random_mutation_search}).

\textbf{Ant Locomotion}:\\
The Ant Locomotion task inherits the robotic body of the Ant task which originally appeared in (\cite{schulman2015high}). The momentary task reward was modified to provide a positive reward signal proportional to velocity in any direction. This was to allow for an adequate comparison with the lifelong learning setting which restricted movement along the x and y plane, requiring a more general policy.

\textbf{Swimmer}:\\
The Swimmer task was originally used in (\cite{coulom2002reinforcement}). This robot is a three-linked 'swimmer' which attempts to attain movement across a plane by extending and retracting its links, much like a snake or worm despite the deceptive environment name. In this task, much like the Ant Locomotion task, reward is based proportionally to velocity along the plane in any direction.

\textbf{CartPole-Swingup}:
The CartPole-Swingup task requires swinging up and balancing a pole on a cart. Reward is proportional to both the cart's distance from the center as well as the pole's distance from being upright. This task requires much more fidelity and consistency than do the locomotion tasks.

\end{document}

%% file: math_commands.tex
\usepackage{amsmath,amsfonts,bm}

\def\eqref#1{equation~\ref{#1}}

\def\1{\bm{1}}

\DeclareMathAlphabet{\mathsfit}{\encodingdefault}{\sfdefault}{m}{sl}
\SetMathAlphabet{\mathsfit}{bold}{\encodingdefault}{\sfdefault}{bx}{n}

